\documentclass[10pt,twocolumn,letterpaper]{article}

\usepackage{cvpr}              %

\newcommand{\ourmethod}{LA-Pose\xspace}
\newcommand\mypara[1]{\noindent\textbf{#1.}}

\usepackage[accsupp]{axessibility} 
\usepackage{multirow}
\usepackage{makecell}

\definecolor{cvprblue}{rgb}{0.21,0.49,0.74}
\usepackage[pagebackref,breaklinks,colorlinks,allcolors=cvprblue]{hyperref}
\usepackage{caption}

\def\paperID{xxxxxx} %
\def\confName{CVPR}
\def\confYear{2026}

\title{LA-Pose: Latent Action Pretraining Meets Pose Estimation}

\author{
Zhengqing Wang\textsuperscript{1,2*} \quad
Saurabh Nair\textsuperscript{1*} \quad
Prajwal Chidananda\textsuperscript{1*} \quad \\
Pujith Kachana\textsuperscript{1} \quad 
Samuel Li\textsuperscript{1} \quad
Matthew Brown\textsuperscript{1} \quad
Yasutaka Furukawa\textsuperscript{1,2} \\[6pt]
\textsuperscript{1}Wayve \quad
\textsuperscript{2}Simon Fraser University
}

\usepackage{graphicx}
\usepackage{caption} %

\begin{document}
\twocolumn[{%
\renewcommand\twocolumn[1][]{#1}%
\maketitle

\includegraphics[width=1.0\textwidth]{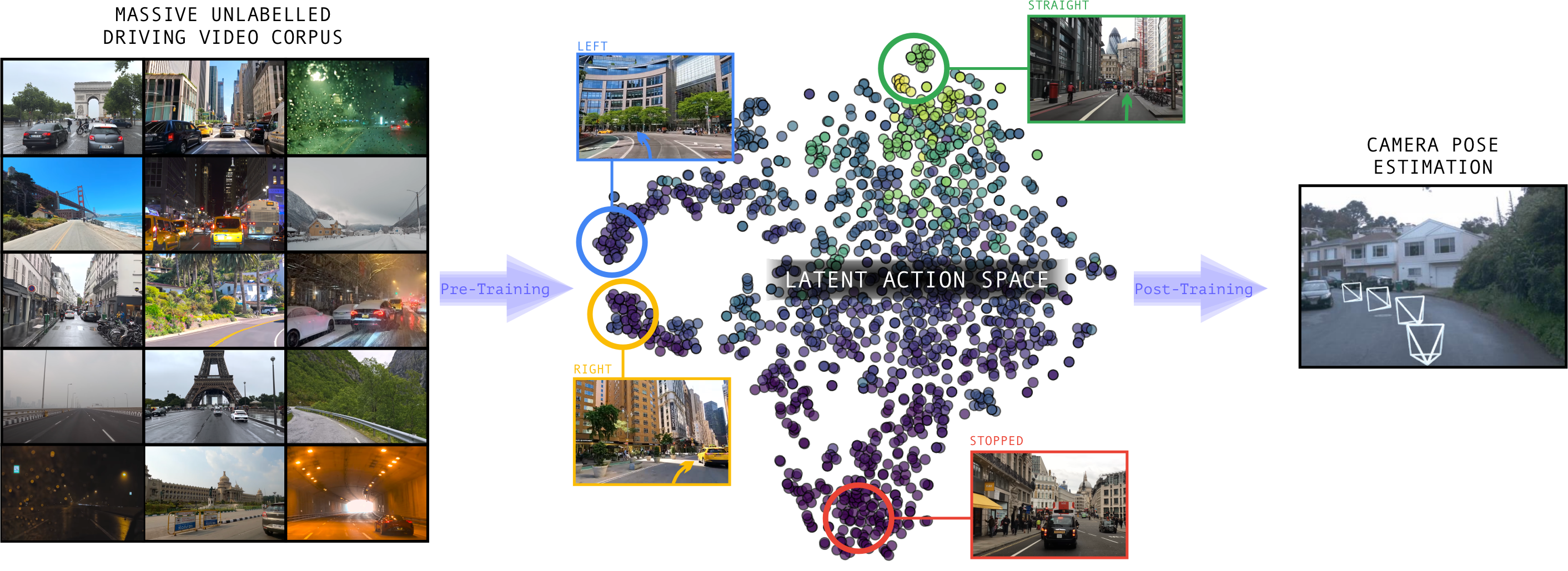}
\vspace{-2em}
\captionof{figure}{
\textbf{Overview of \ourmethod.} 
We introduce a two-stage framework that unifies large-scale \textit{latent action pretraining} with \textit{camera pose estimation}. From millions of unlabeled driving videos,
an inverse--forward dynamics model learns \textit{latent actions} that encode inter-frame motion in a fully self-supervised manner. When visualized in T-SNE space, these latent actions exhibit structured clusters that align closely with true ego-motion distributions. We then re-purpose these representations through lightweight supervised post-training on limited 3D-annotated data, enabling feed-forward pose prediction that is both accurate and highly generalizable. \textbf{LA-Pose} achieves state-of-the-art pose estimation while requiring \textit{orders of magnitude fewer labeled samples}, demonstrating the power of self-supervised latent action learning for scalable 3D perception.
\vspace{2em}
}
\label{fig:teaser_v2}
}]

\begingroup
\renewcommand\thefootnote{\fnsymbol{footnote}} %
\footnotetext[1]{Equal contribution.}
\endgroup

\begin{abstract}
This paper revisits camera pose estimation through the lens of self-supervised pretraining, focusing on inverse-dynamics pretraining as a scalable alternative to the current trend of fully supervised training with 3D annotations. 
Concretely, we employ inverse- and forward-dynamics models to learn latent action representations, similar to Genie \cite{bruce2024genie} from large-scale driving videos.
Our idea is simple yet effective. Existing methods use latent actions in their original capacity, that is, as ``action'' conditioning of world-models or 
as proxies of robot ``action'' parameters in policy networks.
Our method, dubbed \ourmethod, repurposes the latent action features as inputs to a camera pose estimator, finetuned on a limited set of high-quality 3D annotations.
This formulation enables accurate and generalizable pose prediction while maintaining feed-forward efficiency. Extensive experiments on driving benchmarks show that \ourmethod achieves competitive and even superior performance to state-of-the-art methods while using orders of magnitude less labeled data. 
Concretely, on the Waymo and PandaSet benchmarks, \ourmethod achieves over 10\% higher pose accuracy than recent feed-forward methods.
To our knowledge, this work is the first to demonstrate the power of inverse-dynamics self-supervised learning for pose estimation. 
See the project page for demos and more results: \href{https://la-pose.github.io}{la-pose.github.io}.

\end{abstract}
    
\section{Introduction}
\label{sec:intro}

Internet-scale pretraining has transformed the landscape of AI, notably through Generative Pretrained Transformers (GPTs) spanning text, image, and video domains~\cite{caron2021emerging, assran2023self,carreira2024scaling, achiam2023gpt,assran2025v}. 
Within embodied AI, self-driving stands out as a domain uniquely positioned to ride this wave, fueled by massive collections of driving videos from autonomous fleets and consumer vehicles equipped with dash cameras. For vehicles, motion is a direct consequence of action. Harnessing internet-scale video corpora for vehicle pose estimation—effectively their underlying actions—would be a key technology towards the next generation of self-driving systems.

Concurrently, feed-forward 3D reconstruction techniques such as DUSt3R~\cite{wang2024dust3r}, VGGT~\cite{wang2025vggt}, and Rig3R~\cite{li2025rig3r} are rapidly advancing, achieving impressive accuracy by directly predicting structure and camera poses in a single forward pass. Their success, however, relies on 3D annotations derived from Structure-from-Motion, LiDAR, or simulation engines. High-quality labels are available only in curated datasets that require costly hardware and meticulous calibration, and remain small compared to the vast amount of unlabeled driving video available online. As supervised data has become the bottleneck, surprisingly little effort has been devoted to exploring the potential of self-supervised pretraining—the paradigm that has driven massive breakthroughs in the other domains—for geometry perception tasks such as camera pose estimation studied in this paper.

Our method, dubbed \ourmethod, employs a Genie-style architecture in which an inverse-dynamics module learns latent action representations and a forward-dynamics module uses a latent action to predict the next frame. The latent action model is trained in a fully self-supervised manner, offering strong potential for internet-scale pretraining.
Unlike existing methods which use learned latent action representations in their original capacity, for example, as an ``action'' conditioning of world-models in interactive games~\cite{bruce2024genie} or as proxies of robot ``action'' parameters in policy networks~\cite{ye2024latent}, we focus on leveraging latent actions for understanding ego-motion. In self-driving, actions directly manifest as vehicle motion. Consequently, latent actions that model the transition between consecutive frames inherently encode motion change, capturing a compressed representation of pose. \ourmethod 
repurposes latent actions as inputs to a lightweight pose-estimation head post-trained on a limited set of high-quality 3D annotations. This two-stage design unifies large-scale self-supervised video learning with efficient feed-forward pose prediction.

We evaluate our approach on the Waymo~\cite{sun2020scalability} and PandaSet~\cite{xiao2021pandaset} driving benchmarks, where it outperforms state-of-the-art feed-forward 3D reconstruction methods. Extensive quantitative and qualitative analyses demonstrate that our method achieves the highest pose accuracy while requiring significantly less ground-truth 3D annotation. 
Concretely, on both benchmarks, \ourmethod achieves over 10\% higher pose accuracy than recent feed-forward approaches.
To our knowledge, this work is the first to demonstrate the power of inverse-dynamics self-supervised learning for pose estimation. We hope that the paper encourages more research in self-supervised learning of geometry perception tasks, towards Internet level scaling.

\section{Related Work}
\label{sec:formatting}

\mypara{Camera Pose Estimation} Classical SfM/VO systems (e.g., COLMAP \cite{colmap}) remain the default way to obtain camera trajectories for internet imagery. Initial learning-based approaches attempt to directly regress pose using techniques such as convolutions and diffusion \cite{kendall2016posenetconvolutionalnetworkrealtime, wang2024posediffusionsolvingposeestimation, zhang2024camerasraysposeestimation}. Recent learning-based methods such as DUSt3R \cite{wang2024dust3r} directly output aligned dense 3D pointmaps across images, enabling the recovery of relative camera poses. Subsequent works~\cite{wang2025continuous, wang2025vggt, keetha2025mapanything, Yang_2025_Fast3R, wang2025pi3} remove DUSt3R's post-inference optimizations and use a feed forward method to directly regress multiple camera poses, actively pushing the state-of-the-art performance. Concurrently, Rig3R~\cite{li2025rig3r} uses a dense supervision by predicting pose raymaps to recover the camera poses. These approaches achieve strong results, but are typically bound by their training distribution of labeled 3D data, thus inheriting both their costs and biases.  Instead, \ourmethod targets the same feed‑forward test‑time simplicity while efficiently leveraging large-scale, unlabeled data specifically for pose estimation.

\mypara{Self Supervised Learning}
Recent advances in self-supervised learning have enabled rich, transferable representations across modalities: in images, the seminal DINO~\cite{caron2021emerging} framework showed that Vision Transformers~\cite{dosovitskiy2021imageworth16x16words} can learn semantic features via self-distillation without labels; in videos, the Video-MAE\cite{tong2022videomaemaskedautoencodersdataefficient}, and V‑JEPA~\cite{bardes2024revisiting,assran2025v} approaches extended that to spatiotemporal sequences by predicting masked tokens and capturing dynamics; in language, large-scale autoregressive models like GPT demonstrated emergent reasoning capabilities through unsupervised pretraining on massive image-video corpora~\cite{achiam2023gpt}. More recently, the Scaling 4D Representations~\cite{carreira2024scaling} explored how masked-autoencoding in large-scale video transformers (up to 22B parameters) improves performance on non-semantic 3D + temporal (“4D”) tasks such as depth estimation and camera pose. Some works have similarly explored self-supervised methods for pose understanding \cite{Zhou_2017_CVPR, jiang2025rayzer, mitchel2025trueselfsupervisednovelview}; however, these approaches do not leverage large-scale data for pretraining and primarily emphasize photometric reconstruction accuracy, as they are generally designed for novel-view synthesis tasks rather than accurate camera pose estimation. Building on this trajectory, our work brings the predictive self-supervision paradigm into camera pose estimation, proposing a self-supervised pretraining stage that learns geometry-aware representations without explicit 3D labels and thereby bridges the gap between temporal predictive modeling to accurate ego pose estimation.

\mypara{Latent Action Representation}
Latent action representation has recently emerged as a powerful self-supervised learning paradigm for modeling controllable dynamics.
Genie-1~\cite{bruce2024genie} introduced the idea of inferring latent “actions” that transform past frames into future ones, learning a compact discrete codebook through an inverse-dynamics bottleneck.
The resulting latent actions enable controllable video generation, revealing how temporal transitions can be factorized from visual appearance.
Follow-up frameworks extend this idea to robotics, treating latent-action learning as a self-supervised pretraining stage from unlabeled videos, followed by fine-tuning with action-labeled data for output-action prediction and control~\cite{ye2024latent,tharwat2025latent,cui2024dynamo,gao2025adaworld, schmidt2024learningactactions}.
Our work explores this concept from a different angle—leveraging latent action representations not for generation or reinforcement learning, but as a way to structure self-supervised learning of ego-motion.

\mypara{World Models for Video Prediction}
World models aim to predict future observations by modeling the dynamics of the environment. 
In video prediction, a world model learns the temporal evolution of visual scenes, often conditioned on actions that describe how the state changes over time~\cite{hu2023gaia, russell2025gaia, hassan2024gemgeneralizableegovisionmultimodal, nvidia2025cosmosworldfoundationmodel}. 
Classical formulations rely on explicit action labels, which are available in controlled simulation or robotics settings but not in large-scale video data, limiting their scalability to Internet-scale corpora. 
Recent work bridges this gap by pretraining large world models on unlabeled videos to capture general scene dynamics, and then fine-tuning them with a small amount of action-labeled data to recover explicit control capability~\cite{yu2025gamefactory, gao2024vista, feng2024matrix}. 
However, such approaches still depend on ground-truth actions for adaptation. 
Genie overcomes this limitation by introducing latent actions—discrete representations inferred through inverse dynamics—that replace explicit controls as conditioning signals for the forward predictor. 
This formulation allows a world model to learn structured visual dynamics directly from raw videos without requiring action supervision, enabling scalable training on Internet-scale video collections.

\section{LA-Pose}

\begin{figure*}[t]
\centering
\includegraphics[width=\textwidth]{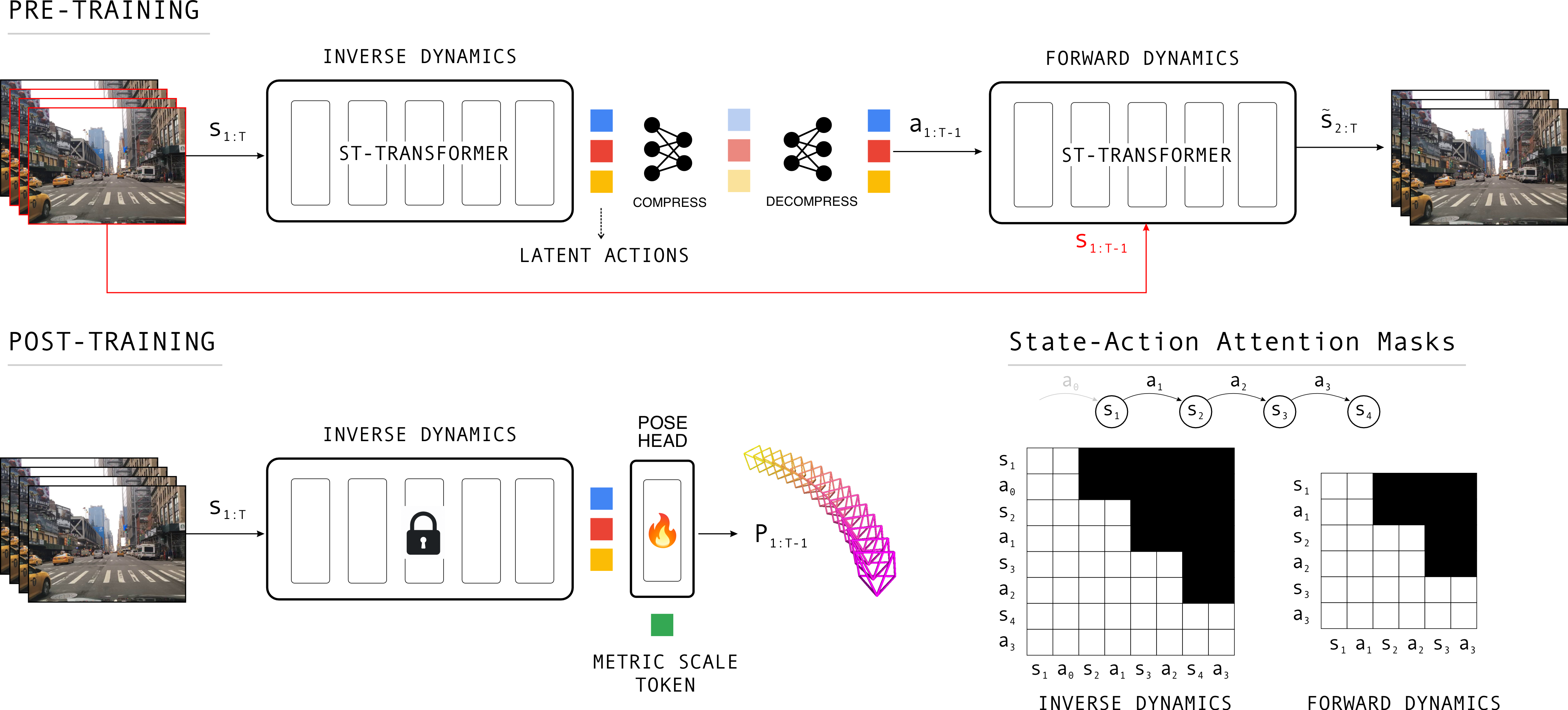}
\caption{Our framework consists of two stages: latent action pretraining and camera pose post-training.
In the pretraining stage (top), an inverse–forward dynamics model learns latent actions from consecutive video frames by predicting future tokens through a self-supervised inverse-dynamics objective. These latent actions encode compact, motion-centric representations of frame-to-frame dynamics.
In the post-training stage (bottom left), we attach a lightweight pose estimation head to the pretrained inverse dynamics encoder. The head predicts relative camera translation, rotation (quaternion), field-of-view, and metric scale from the latent actions.}
\label{fig:lapose_architecture}
\end{figure*}

\ourmethod has two stages of training: latent action pretraining and camera pose post-training, as shown in \autoref{fig:lapose_architecture}. Latent action pretraining is built upon the Genie architecture~\cite{bruce2024genie} with several simplifications tailored to our goal of pose estimation.
Camera pose post-training learns a lightweight pose estimation head that takes the latent actions from the pretrained inverse dynamics model and estimates relative camera poses and metric scale.

\subsection{Latent Action Pretraining}
The latent action model of Genie~\cite{bruce2024genie} includes an inverse dynamics model and a forward dynamics model, which we adopt to learn latent actions for pose estimation. Genie also has a video tokenizer and an additional forward dynamics model. However, these two components are for high-quality video generation and are not used in our work. 
This section provides high level description of the Genie architecture and focuses on our modifications.

\mypara{Image Tokenizer}
An input to the system is a sequence of $T$(=16) frames ($X_1, X_2, \ldots X_{T}$).
The image resolution is 960$\times$448 during training (See \autoref{sec:experiments} for inference and more details). Each image $X_t$ is tokenized into a set of visual tokens, collectively denoted as $s_t$. Thus, the sequence of image states is represented as \{$s_1, s_2, \ldots, s_{T}$\}.
We follow the standard Vision Transformer design, where patch tokens are added with learnable positional embeddings to preserve spatial layout. To encode temporal information, we use sinusoidal frequency-based temporal embeddings, which are projected through an MLP to match the token dimension. This design allows the model to handle variable frame rates and temporal gaps naturally. 
The tokens are processed by a 12-layer transformer encoder, resulting in a 15$\times$7$\times$1536 tensor per frame, where 1536 is the feature dimension.

\mypara{Inverse Dynamics Model}
We build on the inverse dynamics model from the Genie architecture, where an ST-Transformer encoder with causal temporal masking produces a sequence of latent actions with one modification. We introduce a 1536-dimensional learnable query token, repeated at all the frames. The query tokens form a set 
$\{\mathbf{q}_1, ..., \mathbf{q}_{T-1}\}$, which become latent action tokens $\{\mathbf{a}_1, \mathbf{a}_2, \ldots \mathbf{a}_{T-1}\}$ at the output of this component. Each query token aggregates information between consecutive frames, serving as the latent action proxy. The causal masking applies to the query tokens in a slightly different way: $a_t$ interacts with frames up to $t+1$.
This module handles the tensor dimension as (16$\times$15$\times$7 + 15)$\times$1536. 16 is the number of frames and '+15' denotes the number of query tokens.
We further include a pair of three-layer MLPs that compress and de-compress the latent action dimension from 1536 to 50 back to 1536, producing a ``compressed version'' at the bottleneck.
We will evaluate the effects of the compression in \autoref{tab:latent-ablation}, while the uncompressed version before the MLPs remains our default for the pose estimation stage.

\mypara{Forward Dynamics Model} Our forward dynamics model also borrows Genie’s ST-Transformer for predicting future frames with two modifications. First, we replace the final MLP head with a lightweight four transformer blocks (each with 4 layers of self-attention and feed-forward) that operate on the decoder states and project to the outputs.
Second, we use a pretrained VQ-VAE codebook as the prediction target.
Specifically, the ground-truth future frames are first encoded by a frozen VQ-VAE encoder into discrete latent codes. Our model predicts the logits over the same codebook for the next frame.

\subsection{Camera Pose Post-training}
Latent action pretraining captures motion changes and supports effective estimation of camera parameters and metric scale.
To facilitate this transfer, we discard the forward dynamics model and attach a relative pose estimation head to the inverse dynamics model. The inverse dynamics model is either frozen or fine-tuned. A small set of training samples with high-quality ground-truth camera poses are used in this post-training step.

\mypara{Relative Pose Representation}
Latent action encodes relative motions from the past frames to the current. Disentangling a metric scale factor from a scale-agnostic representation has been shown to be effective~\cite{wang2025moge, keetha2025mapanything}. Therefore, the pose head outputs a metric scale factor through an additional token, alongside poses with scale-agnostic translation components.
Specifically, given ground-truth metric motion parameters, we first convert to metric relative motions, where $\{\mathbf{t}_1, \dots, \mathbf{t}_T\}$ denote per-frame translation components.
We compute the average translation magnitude as the metric scale:
$s = \text{mean}_i(\|\mathbf{t}_i\|_2)$. 
We obtain scale-agnostic relative motions by normalizing the translations $\tilde{\mathbf{t}}_i = \mathbf{t}_i / \max(s, \epsilon)$. $\epsilon$ is set to 1.0 for numerical stability. Note that we vary the frame rate during training (\autoref{sec:experiments}) and perform this normalization per given frame rate.

\mypara{Pose Estimation Head} 
The pose head introduces a single learnable metric scale token ($\mathbb{R}^{\text{1536}}$). Together with the latent action tokens $\mathbb{R}^{\text{15}\times \text{1536}}$, a transformer with self-attention (non-causal) processes all tokens, allowing the metric scale token to aggregate information across the sequence. 
The decoder outputs are passed through separate MLP heads: one predicts the 7D relative pose (3D translation, 4D quaternion rotation) with the 1D field-of-view, and another predicts a scalar metric scale with exponential activation to ensure positivity and training stability.

\subsection{Training} \label{sec:training}
\mypara{Training Losses}
The pretraining loss is a cross-entropy loss between the predicted logits and the ground-truth code indices. The post-training loss is L1 losses on normalized translation, quaternion rotation, field-of-view, and log-space metric scale, where we either freeze or fine-tune the inverse dynamics component.

\mypara{Training Data}
We pre-train on an internal corpus of 10.2 million unlabeled driving video-snippets collected from diverse online and proprietary sources, covering a wide range of environments, traffic densities, and weather conditions. This large-scale dataset provides motion supervision for the inverse-dynamics objective. 

For post-training, we use a tiny fraction of that scale, training only on a small set of high-quality labeled data from Waymo~\cite{sun2020scalability}, nuScenes~\cite{caesar2020nuscenes}, and Argoverse~\cite{chang2019argoverse}, where accurate LiDAR-calibrated poses provide metric-scale supervision. The Waymo, nuScenes, and Argoverse datasets contain 750, 850, and 700 scenes, respectively, each lasting approximately 20, 20, and 10 seconds. We generate each training sample by selecting a scene and a frame, while randomizing the frame rate between 1fps and 4fps.

A training sample
consists of 16 consecutive frames from a single front-facing camera. To capture a wide range of motion patterns, the frame stride is randomly sampled between 1 fps and 4 fps during pre-training. This temporal jitter encourages the model to learn motion dynamics over both short- and long-term time horizons. 

\mypara{Training Details}
Pre-training is conducted on 32 H100 GPUs with a global batch size of 64 for 160k steps using a cosine learning rate schedule (peak $1\times10^{-4}$, end $4.5\times10^{-5}$, warm-up 1.5k steps).
Post-training is performed on 8 H100 GPUs for 100k steps with a total batch size of 128 using a cosine learning rate schedule (peak $1\times10^{-4}$, decayed to 0 by the end, warm-up 4k steps). Pre-training completes in approximately four days, and post-training in two days. Notably, our computational cost is significantly lower than that of competing methods. For example, VGGT requires 64 A100 GPUs for nine days of training.
These training details correspond to the setup where the pretrained backbone is frozen during post-training, which is our default configuration.
Details of the fine-tuning setup are provided in \autoref{sec:exp_freeze_finetune} together with the ablation study comparing the two configurations.

\begin{figure*}[t]
\centering
\includegraphics[width=\textwidth]{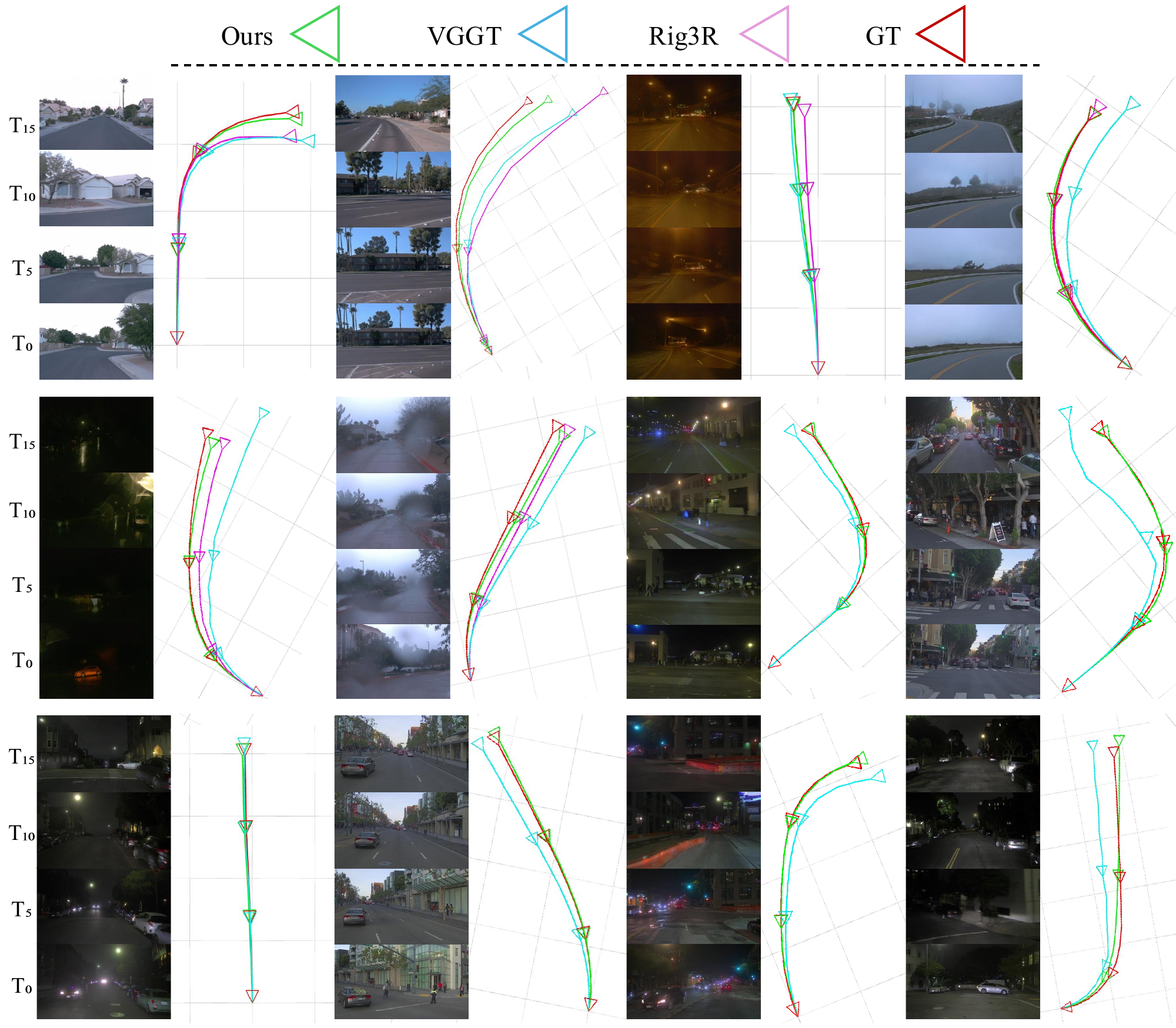}
\caption{Qualitative results of camera pose estimation. Comparison of predicted camera trajectories: Ours (green), Rig3R \cite{li2025rig3r} (magenta), VGGT \cite{wang2025vggt} (cyan), and ground truth (red). Camera frustums are visualized at timestamps 0, 5, 10, and 15, with trajectory lines connecting all camera positions in the sequence. The first six examples show results on the Waymo dataset, while the last six are from PandaSet. All trajectories are projected onto the $xz$ plane for visualization clarity.}
\label{fig:qualitative}
\end{figure*}

\section{Experiments} \label{sec:experiments}

\subsection{Evaluation Setup}
\mypara{Datasets}
We evalaute on the Waymo Open \cite{sun2020scalability} and PandaSet \cite{xiao2021pandaset} datasets. The pre-training stage is performed on an internal driving video corpus, while post-training and evaluation uses Waymo, nuScenes, and Argoverse datasets (See \autoref{sec:training}). Waymo serves as the in-distribution benchmark, where we use the official training and validation split for \ourmethod post-training and evaluation, respectively. 
PandaSet serves as a zero-shot benchmark for all evaluated methods. Both datasets contain large-scale, multi-camera driving scenes with precise LiDAR-based camera pose annotations. 
For evaluation, we uniformly sample frames at 2 fps, corresponding to an 8-second duration over 16 frames).

\mypara{Metrics}
We report three standard metrics: (1) AUC@5, the area under the cumulative error curve up to 5°, computed from pairwise relative rotation and translation angle errors between all frame pairs, reflecting overall pose estimation precision; (2) ATE-S, the scale-invariant aligned trajectory error (RMSE), computed after normalizing trajectories to unit average magnitude and performing global SE(3) alignment, measuring trajectory consistency; and (3) ATE-M, the metric-scale ATE without normalization, reported only when baselines provide metric-scale predictions. When computing scale-invariant metrics, near-stationary sequences (average translation $<$0.1m) are filtered to ensure numerical stability during normalization.

\mypara{Baselines}
We compare with three recent state-of-the-art pose estimation models: Rig3R \cite{li2025rig3r}, VGGT \cite{wang2025vggt}, and MapAnything \cite{keetha2025mapanything}. 
Rig3R is trained on a strictly broader set of driving datasets than \ourmethod. It includes all datasets used in our training (Waymo, nuScenes, and Argoverse) and additionally PandaSet, KITTI, and several other scene datasets with full 3D supervision. Both 
VGGT and MapAnything are trained on Mapillary for driving scenarios, with VGGT also using Virtual KITTI2, but both models include a wide variety of non-driving scene datasets with dense geometric labels. 
Overall, all baseline methods rely on substantially larger amounts of supervised 3D data, whereas \ourmethod combines large-scale self-supervised pre-training with fine-tuning on a limited number of labeled driving sequences.

\subsection{Main Results}

\begin{table}[t]
\centering
\caption{Pose estimation results, reporting area under the curve at an error threshold of 5° \textit{(AUC@5)}, the average aligned trajectory error in scale units \textit{(ATE-S RMSE)}, and in meters \textit{(ATE-M RMSE)}. As Rig3R was trained on the complete PandaSet dataset, it is excluded from the PandaSet evaluation.}
\label{tab:pose_results}

\setlength{\tabcolsep}{4pt}
\renewcommand{\arraystretch}{1.05}

\resizebox{\columnwidth}{!}{%
\begin{tabular}{@{}l@{\hskip 4pt}|ccc@{}}
\toprule
\multicolumn{4}{c}{\textbf{Waymo}} \\
\midrule
\multirow{2}{*}{Method} 
& AUC@5~$\uparrow$ & ATE-S~$\downarrow$ & ATE-M~$\downarrow$ \\
& \% & $\times10^{-2}$ & m \\
\midrule
Rig3R~\cite{li2025rig3r} & \textcolor{orange}{77.9} & 3.17 & - \\
VGGT~\cite{wang2025vggt} & 74.8 & \textcolor{orange}{1.43} & - \\
MapAnything~\cite{keetha2025mapanything} & 65.0 & 3.00 & 4.74 \\
\midrule
\ourmethod & \textcolor{cyan}{91.4} & \textcolor{cyan}{1.20} & \textcolor{cyan}{0.88} \\
\midrule
\multicolumn{4}{c}{\textbf{PandaSet (unseen)}} \\
\midrule
VGGT~\cite{wang2025vggt} & \textcolor{orange}{75.0} & \textcolor{cyan}{0.99} & - \\
MapAnything~\cite{keetha2025mapanything} & 62.4 & 2.75 & 7.28 \\
\midrule
\ourmethod & \textcolor{cyan}{86.3} & \textcolor{orange}{1.13} & \textcolor{cyan}{0.86} \\
\bottomrule
\end{tabular}%
}
\end{table}

\mypara{Quantitative Comparison}
\begin{figure}[h]
\centering
\includegraphics[width=\linewidth]{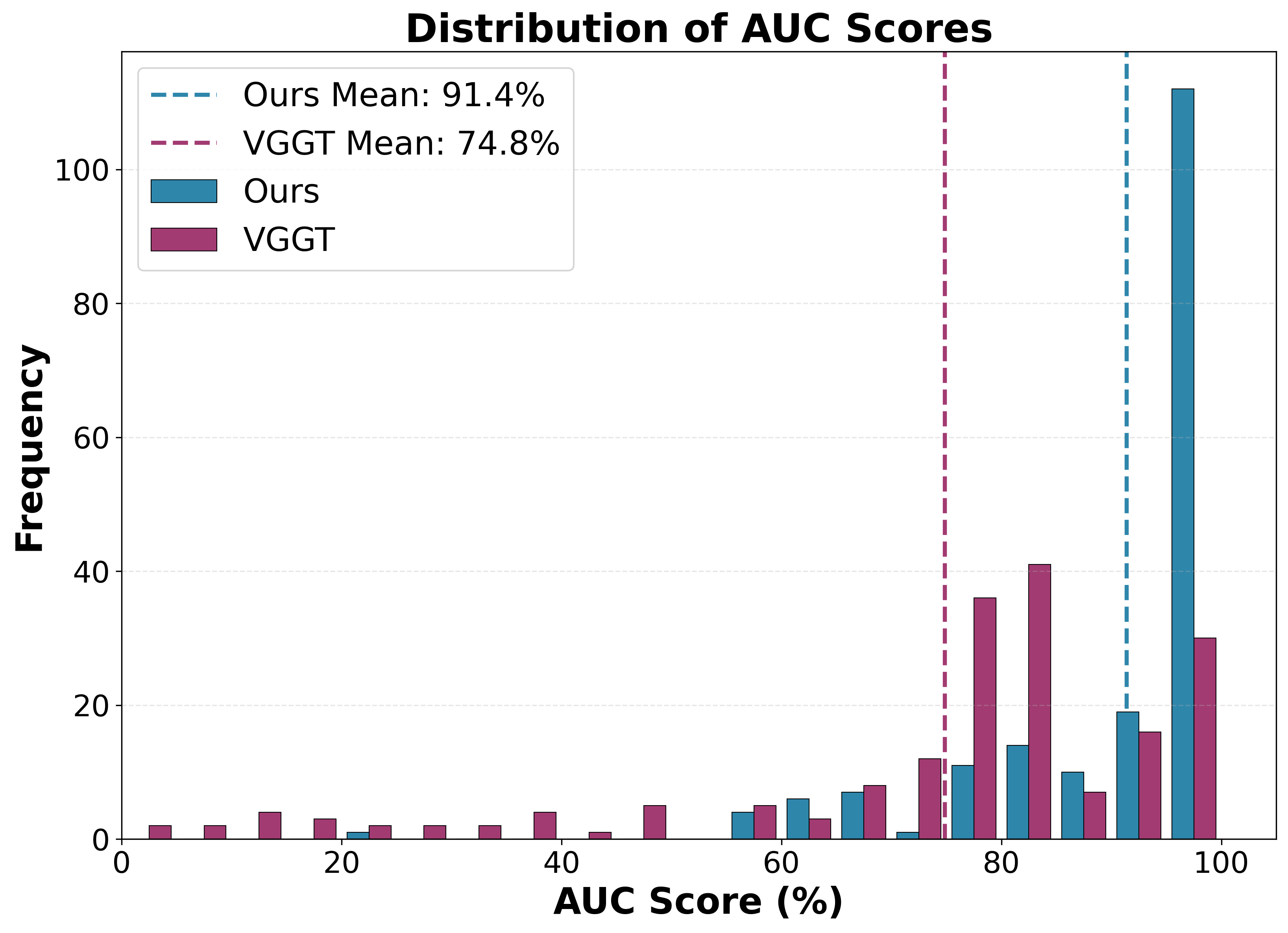}
\caption{
Distribution of pose estimation AUC@5 for \ourmethod and VGGT on the Waymo Open Dataset.
}
\label{fig:auc_histogram}
\end{figure}

~\autoref{tab:pose_results} presents quantitative results on the Waymo and PandaSet benchmarks.
On Waymo, \ourmethod attains an AUC@5 of 91.4\% and an ATE of 1.20$\times10^{-2}$.
On the unseen PandaSet benchmark, our model maintains strong generalization with an AUC@5 of 86.3\%, surpassing all baselines, while the ATE remains comparable to VGGT (1.13$\times10^{-2}$ vs. 0.99$\times10^{-2}$).

\autoref{tab:pose_results} reports the mean of each metric across all samples. To provide a more detailed view, we analyze the distribution of AUC@5 scores in \autoref{fig:auc_histogram}. \ourmethod not only achieves a higher average AUC but also exhibits substantially lower variance across test scenes. Most of our samples cluster near perfect accuracy, while VGGT shows a wider spread with a long tail of low-performing cases. This indicates that \ourmethod delivers more consistent and reliable pose estimation across diverse scenarios, rather than excelling only on easy sequences.

\mypara{Qualitative Analysis}
\autoref{fig:qualitative} visualizes representative examples of estimated camera poses and corresponding images. To provide a balanced view beyond overall averages, we refer to the AUC@5 distribution in \autoref{fig:auc_histogram} and focus our qualitative analysis primarily on the more challenging cases for both \ourmethod and VGGT. From the 185 evaluation samples, we examined those with relatively lower AUC scores and selected a diverse subset capturing different conditions, including rain, nighttime, fog, and sharp turns.
Even in these difficult scenarios, \ourmethod consistently produces stable and geometrically coherent trajectories, highlighting its robustness under adverse conditions. Notably, the post-training data are dominated by simple straight-motion sequences without targeted sampling of rare cases. This robustness emerges naturally from large-scale self-supervised pretraining, which exposes the model to diverse motion patterns and visual appearances.

\subsection{Ablation on freezing/finetuning the backbone} \label{sec:exp_freeze_finetune}

We study the effect of freezing versus fine-tuning the pretrained inverse dynamics model.
~\autoref{fig:ablation_freeze} compares the performance on the Waymo and PandaSet benchmarks.
The x-axis indicates the number of pre-training samples.
Note that a batch size is 64, and 40k post-training steps equal to roughly 2.6 M training samples. 80k steps correspond to 5.2 M.
We report the ATE-M metric.

The figure shows that both configurations achieve comparable performance on the in-domain Waymo benchmark. On the zero-shot PandaSet benchmark, the fine-tuning version degrades significantly, indicating that
freezing the backbone preserves the motion priors learned during pretraining and achieves superior generalization.
\begin{figure}[ht]
  \centering
  \includegraphics[width=\linewidth]{pictures/ablation_freeze_v2.png}
  \caption{
Comparison of pose post-training with frozen and fine-tuned inverse-dynamics encoders. The x-axis shows pretraining data scale. The y-axis reports metric-scale ATE-M (↓). Both settings perform similarly on Waymo, while the frozen backbone generalizes markedly better to the unseen PandaSet.}
  \label{fig:ablation_freeze}
\end{figure}

\subsection{Ablation on Latent Action Dimension}

We study how the dimension of the latent action feature influences both self-supervised pre-training and downstream pose estimation. To save computation, we conduct this ablation using 8 × H100 GPUs with batch size 16 for both pretraining and finetuning. Pre-training was performed over a smaller set 3.2M samples, and post-training was limited to 60k steps.

\autoref{tab:latent-ablation} summarizes the effect of varying the latent action dimensionality. A larger latent space (e.g., 1536-D) achieves lower reconstruction loss during pre-training, as it directly encodes dense motion flow and appearance cues, making the forward dynamics prediction easier. However, this leads to information leakage and weaker abstraction of ego-motion, which is detrimental to pose estimation. In contrast, a smaller latent space (e.g., 50-D) yields higher pre-training loss but promotes compact, motion-centric representations that transfer more effectively to downstream pose estimation. When freezing the pretrained inverse dynamics encoder and training only the pose estimation head, the 50-D latent achieves nearly the same AUC@5 as the 1536-D variant while significantly improving ATE-M, confirming that stronger compression enhances motion awareness and metric-scale consistency.

\begin{table}[t]
\centering
\caption{\textbf{Ablation on latent action dimension.} 
Comparison between different latent dimensions under the \textit{post-training} setting on Waymo. 
Larger latent spaces yield lower reconstruction loss during pre-training but enable information leakage, 
leading to degraded downstream pose estimation. 
Pre-training losses at 100k and 200k steps are measured as cross-entropy on the VQ-VAE code prediction.}
\setlength{\tabcolsep}{5pt}
\renewcommand{\arraystretch}{1.1}
\resizebox{\columnwidth}{!}{%
\begin{tabular}{l|cc|cc}
\toprule
\multirow{2}{*}{\shortstack{Latent\\Dim.}} &
\multicolumn{2}{c|}{Pre-training Loss} &
\multicolumn{2}{c}{Post-training} \\
\cmidrule(lr){2-3}\cmidrule(lr){4-5}
& @100k~$\downarrow$ & @200k~$\downarrow$ 
& AUC@5~$\uparrow$ & ATE-M~$\downarrow$ \\
\midrule
50    & 1.87 & 1.67 & 85.4 & \textcolor{cyan}{1.62} \\
1536  & \textcolor{cyan}{1.35} & \textcolor{cyan}{1.15} & \textcolor{cyan}{86.5} & 1.94 \\
\bottomrule
\end{tabular}%
}
\label{tab:latent-ablation}
\end{table}

\subsection{Robustness to Frame Sampling Rate}

We evaluate the robustness of our model under varying temporal sampling rates, that is, the frame rate of an input video.
A higher fps (smaller temporal gap) provides denser motion cues, while a lower fps corresponds to sparser observations.
As shown in ~\autoref{tab:ablation_fps}, \ourmethod consistently outperforms VGGT~\cite{wang2025vggt} across all fps settings.
While both methods experience a mild decline in performance at lower fps, \ourmethod maintains stable accuracy and low translation error, demonstrating strong temporal robustness.

\begin{table}[t]
\centering
\caption{\textbf{Robustness to frame sampling rate.}
Comparison between \ourmethod and VGGT~\cite{wang2025vggt} under different frame rates on the Waymo benchmark. 
\ourmethod achieves consistently lower trajectory error (ATE-S) and higher pose accuracy (AUC@5).}
\label{tab:ablation_fps}

\begin{tabular}{@{}c|lcc@{}}
\toprule
FPS & Method & AUC@5~$\uparrow$ & ATE-S~$\downarrow$ \\
&   & (\%) & ($\times10^{-2}$) \\
\midrule
4.0     & VGGT & 74.1 & 1.03 \\
                & \ourmethod & \textcolor{cyan}{93.4} & \textcolor{cyan}{0.87} \\
\midrule
1.3    & VGGT & 75.0 & 1.21 \\
                & \ourmethod & \textcolor{cyan}{88.6} & \textcolor{cyan}{1.20} \\
\midrule
1.0    & VGGT & 74.6 & 1.43 \\
                & \ourmethod & \textcolor{cyan}{85.7} & \textcolor{cyan}{1.16} \\
\bottomrule
\end{tabular}%

\end{table}

\section{Limitations, Future work, and Conclusion}

\begin{figure}[t]
\centering
\includegraphics[width=\linewidth]{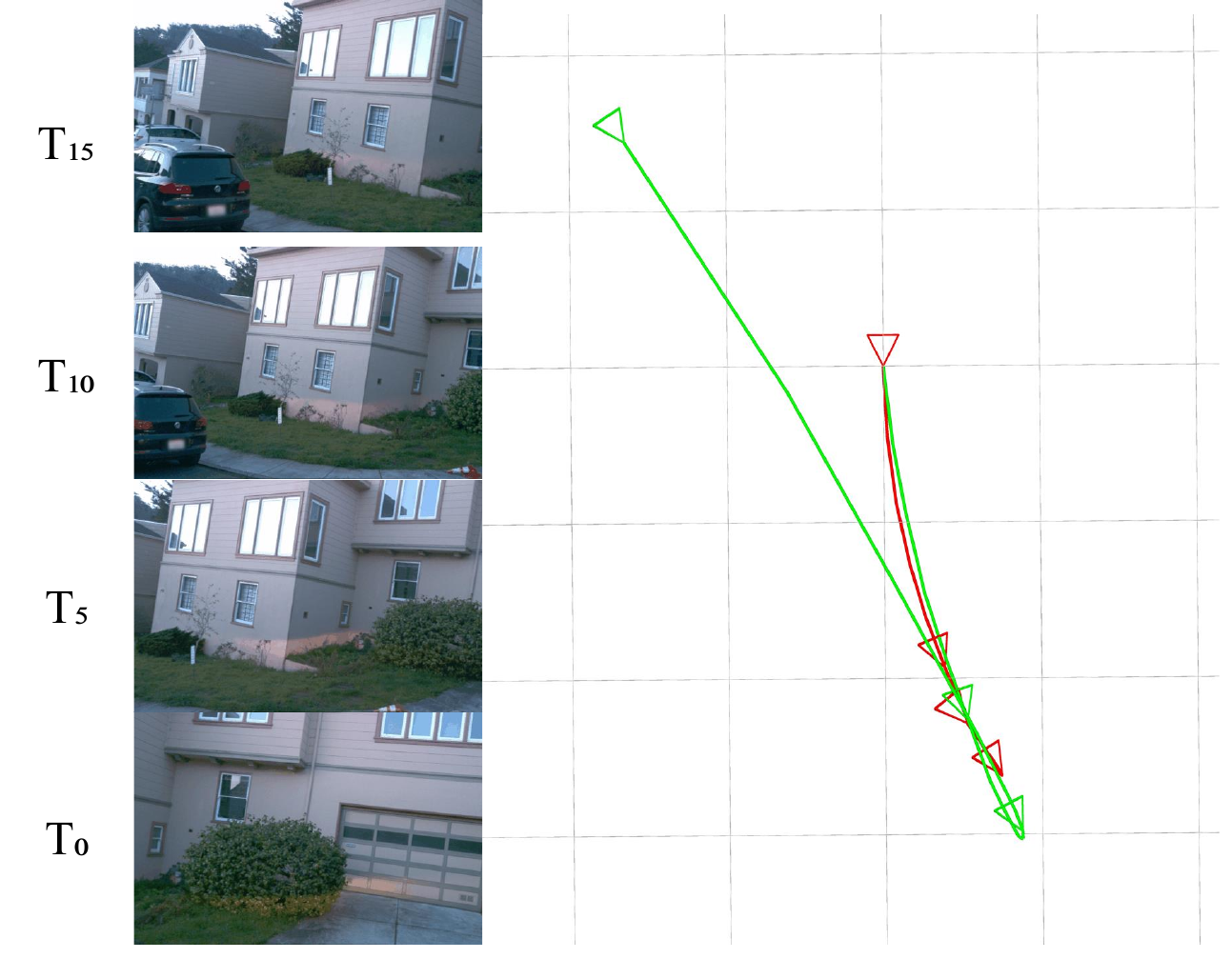}
\caption{\textbf{Failure case under reverse motion.} Performance degrades when the vehicle moves backward, a rare condition in the supervised training set. Despite this distribution gap, the pretrained backbone still produces partially consistent trajectories.
}
\label{fig:failure_case}
\end{figure}

\ourmethod is a self-supervised framework bridging large-scale video pretraining with efficient camera pose estimation. By repurposing latent action features from pretraining as motion-centric features for pose estimation, \ourmethod demonstrates that scalable video pretraining effectively substitutes costly 3D supervision. Experiments on the Waymo and PandaSet benchmarks show that \ourmethod achieves state-of-the-art accuracy with significantly fewer labeled data.
Despite its strong performance, 
we observe degraded accuracy in rare cases such as reverse motion (\ie, backing up), as illustrated in ~\autoref{fig:failure_case}. These scenarios are underrepresented in the supervised post-training data, leading to less stable pose estimates. Future work includes scaling up the pretraining dataset to improve robustness in such rare cases.
More broadly, our formulation of self-supervised video pretraining followed by supervised pose fine-tuning is applicable to other domains. Going beyond driving and extending pretraining to in-the-wild embodied videos--including casual recordings from diverse agents, environments, and camera configurations--could yield general geometric priors transferable across domains.

\section*{Acknowledgments}
\label{sec:acknowledgments}
We thank Jaskaran Singh Sodhi and Saloni Puran Parekh for their help with data construction, and Anner De Jong for engineering support. We are grateful to Thomas Kollar and Gianluca Corrado for reviewing the paper and providing valuable feedback. We also thank Shreyas Rajesh and Soham Phade for their early contributions to the idea of Latent-Actions.

{
    \small
    \bibliographystyle{ieeenat_fullname}
    \bibliography{main}
}

\clearpage
\setcounter{page}{1}
\maketitlesupplementary

\noindent
The supplementary document provides additional qualitative visualizations and analysis:
\begin{itemize}
    \item Qualitative comparison between \ourmethod and VGGT~\cite{wang2025vggt} under the low frame rate (1 fps) setting on Waymo. (\S\ref{supp:low_fps})
    \item Additional qualitative results on the OpenDV–YouTube dataset. (\S\ref{supp:opendv_ty})
    \item Analysis of failure modes across different motion regimes on the Waymo validation set. (\S\ref{supp:failure_analysis})
\end{itemize}

\section{Qualitative Results under Low Frame Rate}
\label{supp:low_fps}

\begin{figure*}[t]
    \centering
    \includegraphics[width=0.85\textwidth]{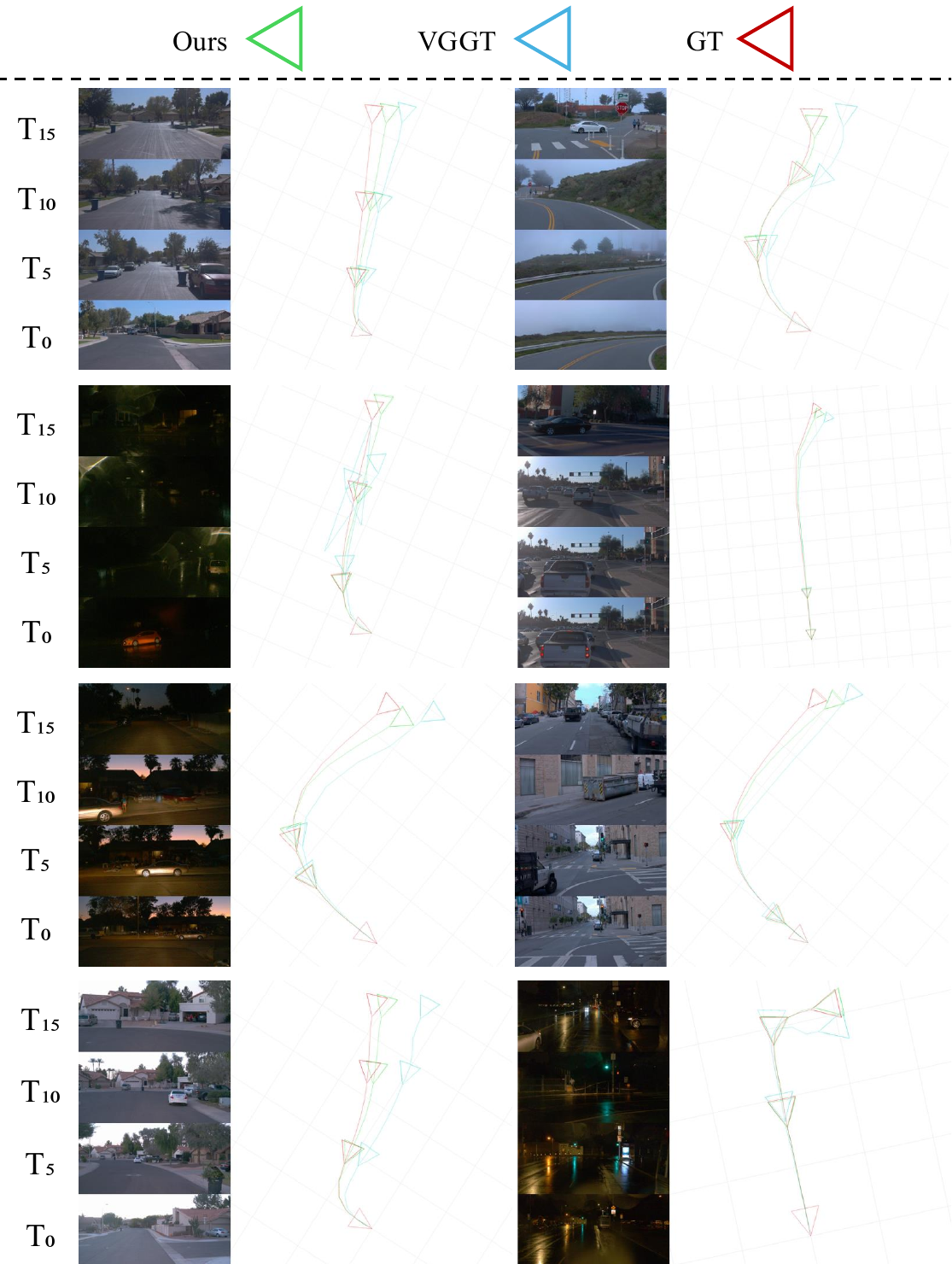}
    \caption{\textbf{Qualitative results under low frame rate (1 fps) on Waymo.}
    Each example shows camera poses projected onto the $xz$ plane, with frustums drawn at frames 0, 5, 10, and 15. 
    \ourmethod (green) maintains stable and temporally consistent motion across the sequence, whereas VGGT~\cite{wang2025vggt} (cyan) exhibits noticeable drift and discontinuities under sparse temporal sampling.}
    
    \label{fig:supp_low_fps}
\end{figure*}

\autoref{fig:supp_low_fps} presents additional qualitative comparisons between \ourmethod and VGGT~\cite{wang2025vggt} under the low frame rate (1 fps) setting on the Waymo dataset. 
All visualizations follow the same protocol as in the main paper, where predicted camera trajectories are projected to the $xz$ plane with camera frustums shown at frames 0, 5, 10, and 15. 

At this extremely sparse temporal sampling, VGGT suffers from noticeable drift and unstable pose transitions, especially along long or turning trajectories. 
In contrast, \ourmethod produces smoother and more consistent camera trajectories, maintaining stable motion even with large temporal gaps between frames. 
These qualitative results highlight the robustness of our learned latent action representation when operating under low frame rate conditions.

\section{Qualitative Results on OpenDV–YouTube}
\label{supp:opendv_ty}
\begin{figure*}[t]
    \centering
    \includegraphics[width=0.88\textwidth]{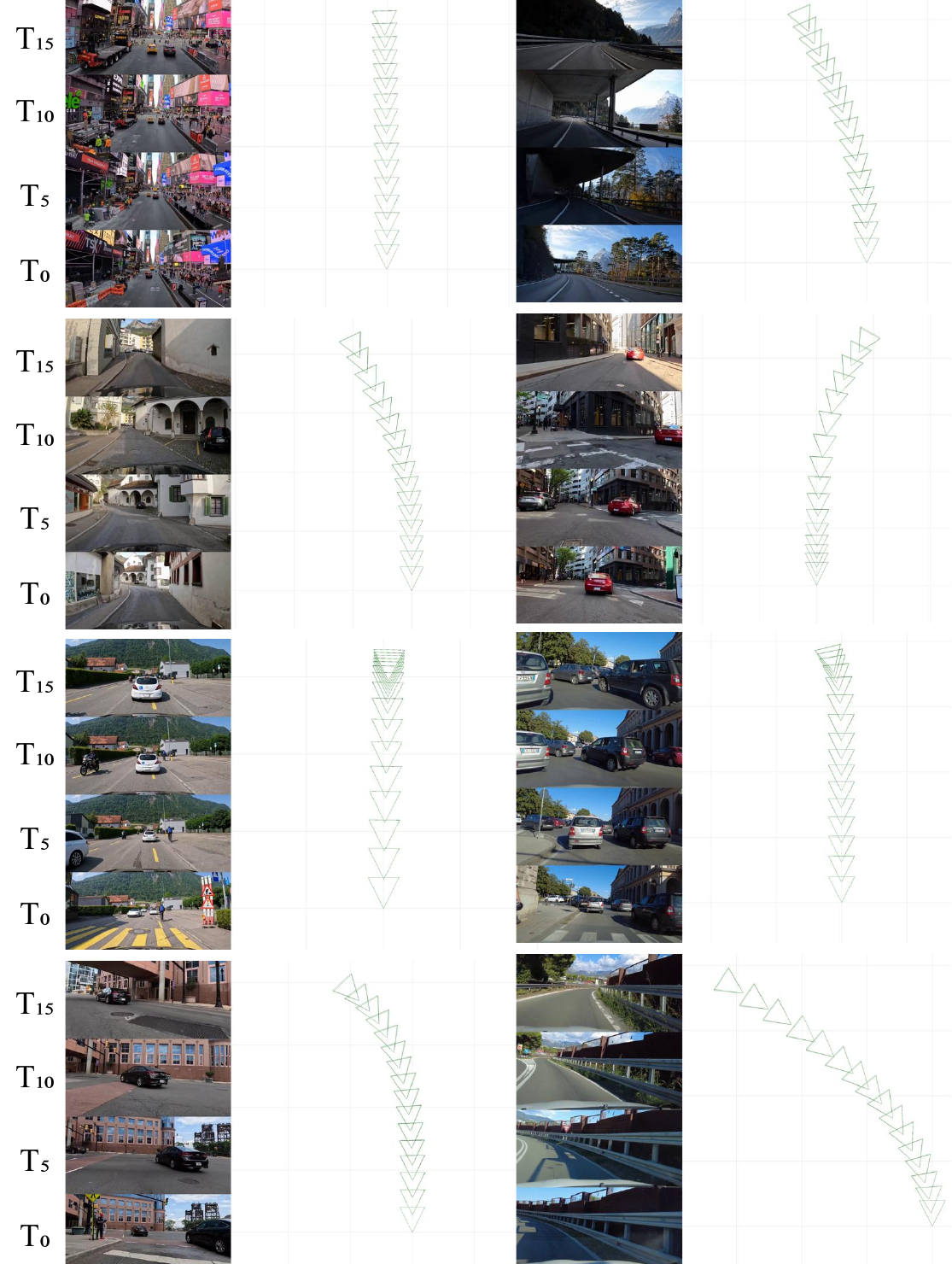}
    \caption{\textbf{Qualitative results on OpenDV–YouTube.}
    Each example shows scenes from diverse cities and viewpoints collected from online YouTube driving videos. 
    \ourmethod produces stable and temporally consistent trajectories across a wide variety of conditions, including urban streets, highways, and curved mountain roads. 
    The results qualitatively demonstrate strong generalization from our pre-trained backbone to uncalibrated, in-the-wild videos.}
    
    \label{fig:supp_opendv_yt}
\end{figure*}

\autoref{fig:supp_opendv_yt} shows qualitative results of \ourmethod on the OpenDV–YouTube dataset~\cite{yang2024genad}. 
The OpenDV–YouTube dataset~\cite{yang2024genad} is a large-scale collection of unconstrained driving videos gathered from public YouTube channels. 
It forms the main component of OpenDV-2K, spanning over 1700 hours of front-view recordings captured across more than 40 countries and 240 cities, far exceeding the geographic coverage of our post-training datasets such as Waymo (San Francisco, Phoenix), nuScenes (Boston and Singapore), and Argoverse (six U.S. cities).This vast diversity covers a wide range of road types, weather conditions, lighting, and camera setups, making OpenDV–YouTube an extremely challenging for evaluating generalization. 

OpenDV–YouTube consists of uncalibrated front-view driving videos collected from YouTube, recorded with diverse in-vehicle cameras of unknown intrinsic and extrinsic parameters. Since the dataset contains only raw RGB frames without ground-truth camera poses, we visualize our predicted camera trajectories to qualitatively assess pose consistency and realism.

We attribute this strong generalization capability to our pre-training stage, which learns a robust video representation from large-scale unlabeled data. 
The frozen backbone, pre-trained on massive driving video corpora, serves as a powerful feature extractor that captures high-level motion patterns and latent action structures. 
This foundation enables the model to transfer effectively to unseen conditions and datasets like OpenDV–YouTube, maintaining stable pose predictions even in unstructured, out-of-distribution environments.

\section{Failure Mode Analysis}
\label{supp:failure_analysis}

To further understand the limitations of our method, we analyze pose estimation performance across different trajectory curvatures and accelerations on the Waymo validation set. 
We categorize trajectories into bins based on curvature and acceleration to examine model performance under different motion regimes. 
Curvature is defined as $\kappa = d\psi/ds$, where $\psi$ denotes the vehicle heading and $s$ is the trajectory arc length. 
We divide curvature into three ranges: small ($<0.01$ m$^{-1}$), medium ($0.01$--$0.1$ m$^{-1}$), and large ($>0.1$ m$^{-1}$). 
Similarly, acceleration magnitude is divided into three ranges: $<0.3$, $0.3$--$0.8$, and $>0.8$ m/s$^2$.

Table~\ref{tab:analysis} reports AUC@5 (\%) under these motion regimes. We observe that performance is lower on medium-curvature trajectories compared to straight or sharp-turn motions. Medium-curvature trajectories correspond to gradual steering behaviors where frame-to-frame geometric changes are subtle. In these situations, visual motion cues between consecutive frames are weak, making the motion representation learned through future-frame prediction less discriminative. By contrast, straight trajectories exhibit stable ego-motion patterns, while sharp turns introduce stronger geometric changes that are easier for the model to capture.

\begin{table}[h]
\centering
\caption{AUC@5 (\%) across different trajectory curvatures and accelerations on the Waymo validation set.}
\label{tab:analysis}
\small
\begin{tabular}{lccc}
\toprule
& Small & Medium & Large \\
\midrule
Curvature & \textcolor{cyan}{94.50} & 78.32 & \textcolor{orange}{91.22} \\
Acceleration & \textcolor{cyan}{92.81} & \textcolor{orange}{90.96} & 88.25 \\
\bottomrule
\end{tabular}
\end{table}

\clearpage

\clearpage

\end{document}